\documentclass[10pt,twocolumn,letterpaper]{article}

\usepackage[pagenumbers]{cvpr} 

\definecolor{cvprblue}{rgb}{0.21,0.49,0.74}
\usepackage[pagebackref,breaklinks,colorlinks,allcolors=cvprblue]{hyperref}

\usepackage{algorithm}
\usepackage[noend]{algpseudocode}
\usepackage{amsfonts}
\usepackage{amsmath}
\usepackage{booktabs}
\usepackage{caption}
\usepackage{enumitem}
\usepackage{hyperref}
\usepackage{makecell}
\usepackage{multirow}
\usepackage{natbib}
\usepackage{wrapfig}
\usepackage{graphicx}
\usepackage{subcaption}
\usepackage{xcolor}
\usepackage{tcolorbox}
\usepackage{lipsum}
\usepackage{soul}
\usepackage{pifont}
\usepackage{appendix}
\usepackage{tocloft} 
\usepackage{hyperref} 
\usepackage{circledsteps}
\usepackage{float}
\usepackage{subfloat}

\usepackage{tcolorbox}
\usepackage{fontawesome}
\usepackage{marvosym}

\usepackage[margin=1in]{geometry}
\usepackage{amssymb}
\usepackage{array}

\def\paperID{10} 
\def\confName{CVPR}
\def\confYear{2025}

\definecolor{amber(sae/ece)}{rgb}{1.0, 0.49, 0.0}

\tcbuselibrary{skins, breakable}

\tcbset{
	myboxstyle/.style={
		breakable,
		colback=white,            
		colframe=black,           
		colbacktitle=gray!15,     
		coltitle=black,           
		fonttitle=\bfseries,      %
		title={#1},               
		enhanced,
		attach boxed title to top left={yshift=-2mm, xshift=2mm},
		boxed title style={
			colframe=black,
			arc=1mm,
			outer arc=1.5mm,
			boxrule=0.5mm,        
		},
		overlay={},
	}
}

\definecolor{maroon}{cmyk}{0,0.87,0.68,0.32}
\definecolor{myyellow}{RGB}{218, 160, 109}
\definecolor{brickred}{rgb}{0.8, 0.25, 0.33}
\definecolor{brandeisblue}{rgb}{0.0, 0.44, 1.0}
\definecolor{applegreen}{rgb}{0.55, 0.71, 0.0}
\definecolor{aogreen}{rgb}{0.0, 0.5, 0.0}

\definecolor{gdmb}{RGB}{47, 114, 173}  
\definecolor{gdmr}{RGB}{199, 100,  38}
\definecolor{gdmg}{RGB}{70, 155, 118}
\definecolor{gdmm}{RGB}{193, 126, 165}
\definecolor{gdmy}{RGB}{239, 227,  98}
\definecolor{gdmc}{RGB}{110, 179, 228}
\definecolor{gdmk}{RGB}{20, 20, 20}
\definecolor{turquoise}{cmyk}{0.65,0,0.1,0.3}
\definecolor{purple}{rgb}{0.65,0,0.65}
\definecolor{dark_green}{rgb}{0, 0.5, 0}
\definecolor{orange}{rgb}{0.8, 0.6, 0.2}
\definecolor{red}{rgb}{0.8, 0.2, 0.2}
\definecolor{darkred}{rgb}{0.6, 0.1, 0.05}
\definecolor{blueish}{rgb}{0.0, 0.3, .6}
\definecolor{light_gray}{rgb}{0.7, 0.7, .7}
\definecolor{pink}{rgb}{1, 0, 1}
\definecolor{greyblue}{rgb}{0.25, 0.25, 1}
\definecolor{orgred}{rgb}{1.0, 0, 0}

\definecolor{rjGreen}{rgb}{0.02,0.7,0.02}

\title{LangCoop: Collaborative Driving with Language}

\author{
Xiangbo Gao$^{1}$,
Yuheng Wu$^{2}$,
Rujia Wang$^{1}$,
Chenxi Liu$^3$,
Yang Zhou$^1$,
Zhengzhong Tu$^{1*}$\\[0.1cm]
$^1$Texas A\&M University, 
$^2$KAIST,
$^3$University of Utah \\
{\tt\small \{xiangbog, tzz\}@tamu.edu}\\
{\small $^*$ Corresponding author}\\
{\small\href{https://xiangbogaobarry.github.io/LangCoop}{\textcolor[HTML]{D42B8F}{https://xiangbogaobarry.github.io/LangCoop}}}
}
\begin{document}
%
%
\newcommand{\red}[1]{{\color{red}#1}}
\newcommand{\todo}[1]{{\color{red}#1}}
\newcommand{\TODO}[1]{\textbf{\color{red}[TODO: #1]}}

\newcommand{\mcot}{M$^3$CoT }

\newcommand{\Mcot}{Mixture Model Modular Chain-of-thought }
\newcommand{\TZZ}[1]{{\color{orange}TZZ: #1}}


\maketitle

\begin{abstract}

Multi-agent collaboration holds great promise for enhancing the safety, reliability, and mobility of autonomous driving systems by enabling information sharing among multiple connected agents. 
However, existing multi-agent communication approaches are hindered by limitations of existing communication media, including high bandwidth demands, agent heterogeneity, and information loss. 
To address these challenges, we introduce \textbf{LangCoop}, a new paradigm for collaborative autonomous driving that leverages natural language as a compact yet expressive medium for inter-agent communication.
LangCoop features two key innovations: Mixture Model Modular Chain-of-thought (M$^3$CoT) for structured zero-shot vision-language reasoning and Natural Language Information Packaging (LangPack) for efficiently packaging information into concise, language-based messages. 
Through extensive experiments conducted in the CARLA simulations, we demonstrate that LangCoop achieves a remarkable 96\% reduction in communication bandwidth ($<$ 2KB per message) compared to image-based communication, while maintaining competitive driving performance in the closed-loop evaluation. Our project page and code are at \href{https://xiangbogaobarry.github.io/LangCoop/}{https://xiangbogaobarry.github.io/LangCoop/}.
%
%

\end{abstract}

\section{Introduction}
\label{sec:intro}



Recent advances in autonomous driving have demonstrated that multi-agent collaboration~\cite{liu2023towards} significantly enhances both safety and efficiency compared to single-vehicle operations, primarily through real-time information sharing and intention communication.
This collaborative approach has become increasingly crucial as autonomous vehicles navigate complex environments where interaction with other traffic participants is inevitable and constant. 
However, the selection of an appropriate collaboration medium remains a critical challenge that has attracted substantial research attention.

A key element of multi-agent collaboration is the medium used for inter-vehicle communication. Researchers have proposed various modalities for exchanging information, including: raw sensor data, neural network features, and downstream task results.
Despite their utility, each of these communication media suffers from one or more critical drawbacks. Specifically, they often:
\underline{\textbf{(1)}} Require high bandwidth, placing a heavy load on communication infrastructures and increasing the risk of latency or packet loss.
\underline{\textbf{(2)}} Fail to accommodate the inherent heterogeneities across agents, which may use different sensor configurations, model architectures, or targeting on different downstream tasks.
\underline{\textbf{(3)}} Lose critical contextual information when data are overly compressed, abstracted, or otherwise transformed into a limited representation.
\underline{\textbf{(4)}} Does not support planning-level or control-level collaboration.

To address these issues, we propose that \textbf{human natural language} can serve as an effective communication medium for multi-agent collaborative driving. Unlike conventional sensor-based or feature-based communications, natural language is inherently flexible and capable of conveying a broad range of contextual and semantic cues, therefore offering additional advantages. \underline{\textbf{First}}, it bridges the gap between machine-readable modalities~\cite{chiu2025v2v} (e.g., numbers, features, embeddings) and human-spoken language, making the reasoning~\cite{yao2023react, jiang2025alphadriveunleashingpowervlms}, communication~\cite{hu2024pragmatic}, negotiation~\cite{cui2025talking}, and decision-making process more transparent. Such transparency benefits research, development, and debugging by enabling human operators to understand and verify the messages being exchanged among autonomous vehicles. \underline{\textbf{Second}}, ongoing research in leveraging LVLMs within autonomous driving has already demonstrated their utility in understanding~\cite{tian2024drivevlm}, reasoning~\cite{xing2025openemma}, decision-making~\cite{xu2024drivegpt4,sima2024drivelm}, and even low-level vehicle control~\cite{cui2024board}. Consequently, natural language collaboration can synergistically exploit the general intelligence of LVLMs to achieve more robust, versatile, and explainable multi-agent collaboration. \underline{\textbf{Third}}, natural language enables high-level fusion or negotiation at the planning and prediction levels, allowing agents---including automated vehicles, human drivers, and road-side units---to communicate intention and decision rationale rather than just perception data. 
This capability simplifies the coordination process, allowing agents to reach mutual understanding and agreements rapidly and clearly, ultimately promoting smoother, safer, and more socially acceptable driving behaviors.
\underline{\textbf{Lastly}}, 
natural language naturally provides scalability and generalization across diverse scenarios and heterogeneous vehicle platforms. Using standardized language-based communication seamlessly integrates autonomous and human-driven vehicles, regardless of sensor suites or underlying technologies. Moreover, natural language communication is inherently model-agnostic, compatible with both open-source (\eg LLAMA~\cite{grattafiori2024llama}, DeepSeek~\cite{guo2025deepseek}) and commercial LLMs (\eg ChatGPT~\cite{chatgpt}, Gemini~\cite{team2023gemini}), enabling easy adoption and interoperability across diverse autonomous vehicle systems.

Another compelling rationale emerges from real-world autonomous driving incidents, such as a case where a Waymo driverless car stopped dead inside a construction zone, causing disruptions and creating hazards~\cite{waymo2023}.
 Such incidents highlight the fundamental limitation of conventional sensor-based communication: it fails to transparently communicate the vehicle’s internal decision-making and reasoning processes to nearby human drivers or traffic controllers. In contrast, an interface that uses natural language as a universal information protocol could explicitly communicate an autonomous vehicle’s internal reasoning and intentions in real-time (e.g., “I’ve stopped due to unclear construction signage”), thereby clarifying otherwise confusing behaviors, reducing driver frustration, and facilitating timely human intervention. 
 Furthermore, such a natural language-based approach allows real-time human-in-the-loop interaction, enabling remote operators or nearby traffic managers to quickly communicate or disengage with the vehicle in intuitive terms (e.g., “Please move slowly to the side”) to promptly resolve ambiguous or problematic situations.

Leveraging these insights, we introduce \textbf{LangCoop}, a novel framework for collaborative autonomous driving that uses natural language as the primary medium for inter-vehicle communication. Our framework consists of three key components: \underline{\textbf{(1)}} a Mixture Model Modular Chain-of-thought (M3CoT) module that structures reasoning into distinct stages for comprehensive scene understanding; \underline{\textbf{(2)}} a Natural Language Information Packaging (LangPack) system that compresses rich semantic information into compact messages; and \underline{\textbf{(3)}} multiple driving signal generation approaches that translate natural language reasoning into actionable controls. Our experimental results in closed-loop evaluations using the Carla simulator~\cite{dosovitskiy2017carla} show that, by using zero-shot LVLMs, LangCoop achieves driving scores of up to 48.8 and route completion rates of up to 90.3\%, significantly outperforming non-collaborative baselines while maintaining exceptional communication efficiency ($<$2 KB). The framework also operates effectively with heterogeneous agent capabilities, demonstrating the viability of natural language as a medium for autonomous vehicle collaboration.

\section{Related Works}

\subsection{LVLMs in Autonomous Driving}

The integration of Vision-Language Large Models (LVLMs) into autonomous driving has enabled a unified approach to perception, reasoning, and decision-making, offering enhanced interpretability and adaptability~\cite{da2025generative, xing2024autotrust, huang2025trustworthiness, luo2025senserag}. Early studies have explored LVLMs for closed-loop driving, where multimodal sensor data is processed alongside natural language instructions to generate vehicle control outputs. Shao \emph{et al.}~\cite{shao2024lmdrive} introduced one of the first LVLM-based end-to-end driving models, while Wang \emph{et al.}~\cite{wang2023drivemlm} focused on translating language instructions into high-level driving commands. Xu \emph{et al.}~\cite{xu2024drivegpt4} and Sima \emph{et al.}~\cite{sima2024drivelm} further emphasized explainability, using question-answering and graph-based reasoning to interpret scene dynamics and decision rationales, making autonomous systems more transparent and human-interpretable. Hwang \emph{et al.}~\cite{hwang2024emma} used LVLMs to directly output the future planning waypoints. Xing \emph{et al.}~\cite{xing2024autotrust} proposed a comprehensive benchmark for evaluating the truthfulness, safety, fairness, security, and generalizability of LVLMs in the autonomous driving scenes.

Beyond perception, LVLMs have demonstrated robustness in out-of-distribution (OOD) scenarios, addressing challenges that conventional deep-learning models struggle with in unseen environments. Wang \emph{et al.}~\cite{wang2024drive} showed that LVLMs could simulate novel situations through latent space editing, improving generalization. Mei \emph{et al.}~\cite{mei2024continuously} introduced a dual-process framework, combining slow but rigorous reasoning from an LVLM with fast real-time execution from a smaller model, mimicking human cognitive processes. Additionally, Dong \emph{et al.}~\cite{dong2024generalizing} and Xing \emph{et al.}~\cite{xing2025openemma} explored zero-shot prompting, demonstrating how LLMs can guide autonomous systems without extensive retraining.

LVLMs also play a pivotal role in multi-agent collaboration and human-centric driving by improving vehicular communication~\cite{wu2025v2x} and personalized decision-making~\cite{da2025generative,wu2025v2x}. Liang \emph{et al.}~\cite{liang2024generative} and Zhang \emph{et al.}~\cite{zhang2024generative} explored how generative AI models enable semantic-rich, context-aware inter-vehicle communication, surpassing traditional bandwidth-intensive numeric exchanges. In personalized driving, Li \emph{et al.}~\cite{li2024large} highlighted that LVLMs improve context understanding and human-like reasoning, while Lan \emph{et al.}~\cite{lan2024traj} and Duan \emph{et al.}~\cite{duan2024prompting} demonstrated their ability to simulate human driving behaviors and dynamically adjust trajectories. As LVLMs continue evolving, their integration into autonomous systems paves the way for more interpretable, adaptable, and collaborative driving solutions that better align with human expectations and real-world challenges.

\subsection{Collaboration Medium in Multi-agent Driving}

Effective collaboration among autonomous agents in multi-agent driving scenarios hinges on the choice of communication medium. Several approaches have been explored, including the exchange of raw sensor data~\cite{gao2018object, chen2019cooper, arnold2020cooperative}, neural network features~\cite{wang2020v2vnet, guo2021coff, bai2022pillargrid, cui2022coopernaut, xu2022v2x, xu2023cobevt, wang2023core, yu2023vehicle, gao2025stamp, wangcocmt, liu2024towards}, and perception results~\cite{melotti2020multimodal, fu2020depth, zeng2020dsdnet, shi2022vips, glaser2023we}. Specifically, raw sensor data (such as LiDAR point clouds or camera images) offers comprehensive environmental perception but demands high communication bandwidth and latency. Meanwhile, neural network features (intermediate embeddings, BEV feature maps, or feature queries) can reduce bandwidth usage yet introduce incompatibility when agents rely on heterogeneous feature extraction networks. Another approach is sharing perception results, such as predicted depth maps~\cite{hu2023collaboration}, object detection outputs~\cite{xu2023bridging}, occupancy grids~\cite{song2024collaborative}, or BEV map segmentations~\cite{xu2023cobevt}. While enumerating all possible perception outputs can strain communication bandwidth, limiting the shared set risks losing critical semantic details.

\begin{figure*}[t]  
    \centering
    \includegraphics[width=\textwidth]{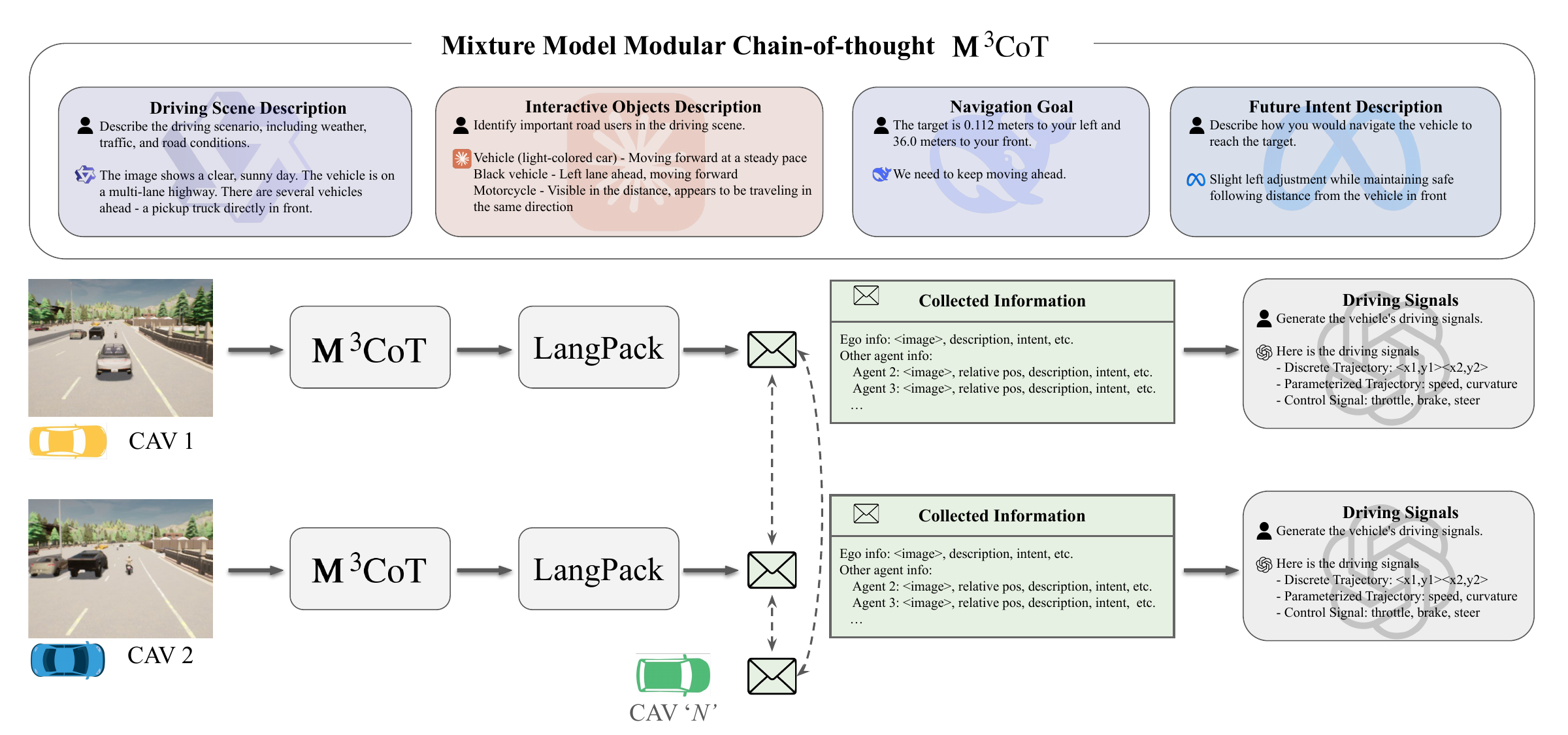} 
    \caption{Overview of the LangCoop framework.}
    \label{fig:overview} 
\end{figure*}

Given these challenges, natural language has emerged as a promising alternative for communication in multi-agent driving. Unlike numeric-based representations, natural language is compact, human-interpretable, and adaptable to heterogeneous agents. It also supports planning or control interactions. Recent studies in robotics and autonomous driving have begun to explore language-based communication, leveraging its ability to capture rich contextual information with minimal overhead.. For instance, Hu \emph{et al.}~\cite{hu2024agentscodriver}, Yao \emph{et al.}~\cite{yao2024comal}, and Fang \emph{et al.}~\cite{fang2024towards} use Large Language Models (LLMs) for driving-scenario reasoning on highly abstract traffic descriptions but overlook pedestrians, cyclists, unknown obstacles, and environmental conditions that are pivotal in real-world driving. Another approach, V2V-LLM ~\cite{chiu2025v2v}, augments an LLM backbone with pre-trained perception features (such as object detections) to incorporate environmental cues. However, it does not exploit the vision-based reasoning capabilities of LVLMs. V2X-VLM~\cite{you2024v2x} is the first work to combine perception and reasoning within a LVLM framework, yet it essentially treats multi-agent collaboration as a multi-sensor fusion problem, neglecting important factors like cross-sensor coordination transformations and collaboration at the planning or control level. Moreover, its evaluation remains limited to open-loop benchmarks, and its model is not open-sourced.

In this work, we advance the field by harnessing both the perception and reasoning capabilities of LVLMs, enabling planning- and control-level collaboration among autonomous vehicular agents. Unlike previous approaches, we conduct closed-loop evaluations to assess real-time performance and provide open-source code for the research community to facilitate further exploration and benchmarking.

\section{Methodology}

\subsection{Framework Overview}

In this section, we present LangCoop, a novel framework that natively leverages Large Vision Language Models (LVLMs) for collaborative driving among Connected Autonomous Vehicles (CAVs). As illustrated in Fig.~\ref{fig:overview}, our framework establishes a systematic pipeline for information extraction, processing, exchange, and decision-making in collaborative driving scenarios. Each CAV initially captures front-view images through its onboard cameras, which serve as the primary sensory input. These images are passed through our \Mcot (\mcot) module (detailed in Section~\ref{sec:prompting_method}), which systematically extracts environmental and object-level information as well as process goal-oriented information, and behavioral intentions. 

The extracted information is then packaged into a compact, structured natural language format via our Natural Language Information Packaging (LangPack) module. This standardized format facilitates information exchange between connected vehicles while minimizing bandwidth requirements. Concurrently, each vehicle receives packets from other CAVs within the communication range. Upon receiving the packets, each vehicle integrates the messages with its own and feeds them into the LVLMs to generate appropriate driving signals. The driving signals are formulated as discrete trajectories, continuous trajectories, or direct control commands depending on the specific implementation context (detailed in Section~\ref{sec:driving_signal}). These signals guide the vehicle's planning and control systems to execute safe and efficient maneuvers.

\subsection{\Mcot}
\label{sec:prompting_method}

The \Mcot (\mcot) module forms the cognitive foundation of our LangCoop framework, expanding upon the chain-of-thought reasoning process introduced by OpenEmma~\cite{xing2025openemma}. \mcot systematically decomposes the complex task of driving scene understanding into four distinct prompting stages, each addressing a specific aspect of the driving context: \textbf{driving scene description} that focuses on holistic environmental understanding, \textbf{interactive object description} that identifies and characterizes specific objects relevant to the driving task, \textbf{navigation goal prompting} that informs the agent about its next navigational goal's relative location, shifting the agent's perspective from mere trajectory prediction to goal-oriented planning, and finally \textbf{future intent description} that articulates the vehicle's intended actions and decision rationale.

A key innovation in our approach is the flexibility to employ different specialized LVLMs for each prompting stage. This design choice offers several significant advantages: First, it acknowledges that different prompting tasks demand distinct capabilities—driving scene and object description rely predominantly on visual understanding capabilities, while navigation goal interpretation and future intent formulation necessitate stronger logical reasoning skills. By selecting models optimized for these specific competencies, our system potentially outperforms monolithic approaches that use a single model for all tasks. Second, this modular design offers practical benefits related to computational efficiency and cost management. Given that zero-shot LVLM inference can be resource-intensive, particularly for high-performance models, our approach allows for strategic resource allocation—deploying more powerful (and potentially more expensive) models only for the stages that critically require their capabilities. This alleviates the need for a single large generalist model, potentially reducing inference time and operational costs without compromising performance.

\subsection{Natural Language Information Packaging}

Our framework introduces Natural Language Information Packaging (LangPack) as an innovative medium for information sharing. LangPack gathers diverse information sources into a standardized, human-readable, and machine-processable format that balances comprehensiveness with transmission efficiency. Upon completing the \mcot processing stages, each vehicle constructs a LangPack packet that integrates prompting results with agent metadata, including location, velocity, acceleration, etc.

\begin{table}[h!]
\caption{Natural Language Information Packaging Structure.}
\centering
\renewcommand{\arraystretch}{1.3}
\small
\begin{tabular}{|p{0.9\linewidth}|}
\hline
\multirow{1}{*}{{\Letter}\ \textbf{Natural Language Information Packaging}}\\
\hline
\textbf{Agent Metadata:} \textit{location, velocity, acceleration, etc.}\\
\textbf{Scene Description:} \textit{The image shows} \dots \\
\textbf{Objects Description:} \textit{Vehicle (light-colored car) - Moving forward \dots} \\
\textbf{Navigation Goal:} \textit{We need to keep moving ahead \dots} \\
\textbf{Intent Description:} \textit{Slight left adjustment while maintaining safe \dots} \\
\dots \\
\hline
Total Package Size: \textbf{$<$ 2KB} \\
\hline
\end{tabular}
\end{table}

The LangPack approach offers several distinct advantages for collaborative driving systems. First, the inherent compactness of natural language representation allows for information-dense communication with minimal bandwidth requirements—typical LangPack packages require less than 2KB of data, making them suitable for transmission even in bandwidth-constrained V2X communication environments. Furthermore, natural language provides a flexible and extensible medium that can accommodate diverse information types without requiring rigid structural redesigns. This adaptability is particularly valuable for autonomous driving systems that must process heterogeneous and sometimes unexpected environmental elements.

Upon receiving LangPack packages from other connected vehicles, each CAV performs essential post-processing operations including coordinate transformation and temporal alignment. The processed information is then aggregated with the vehicle's own perceptions and prompting results to create a comprehensive knowledge ready to be passed into the following decision-making module.

\subsection{Driving Signal Generation}
\label{sec:driving_signal}

The final component of our LangCoop framework involves translating the aggregated, multi-vehicle understanding into actionable driving signals. We propose three driving signal formulations, each offering specific advantages depending on the implementation context and downstream control requirements:

\noindent \textbf{Discrete Trajectory Generation}: The LVLM outputs a sequence of waypoints $(x_i, y_i)$ for the future $n$ seconds. This high-precision path representation is suitable for complex maneuvers and enables straightforward validation against environmental boundaries.

\noindent \textbf{Continuous Trajectory Generation}: Rather than discrete positions, this approach defines vehicle motion through speed and turning curvature parameters over time. It produces smoother motion profiles that better align with vehicle dynamics for natural-feeling behavior.

\noindent \textbf{Direct Control Signal Generation}: In this most direct formulation, the LVLM outputs low-level control signals—specifically steering angle, throttle position, and brake pressure—for each time step. A key advantage of this approach is that outputs can be explicitly constrained within physically feasible ranges (e.g., steering angle limits, maximum acceleration rates), ensuring generated commands never exceed the vehicle's operational capabilities.

\noindent In Section~\ref{sec:signal}, we present a comparative analysis of all three driving signal formulations across diverse driving scenarios.

\section{Experiments}
\label{sec:exp}

In this section, we present comprehensive experimental evaluations of our LangCoop framework through closed-loop simulations in the CARLA environment~\cite{dosovitskiy2017carla}. We first outline our experimental setup and evaluation metrics (\S~\ref{sec:setup}), followed by a systematic assessment of key components within our framework, including driving signal formulations (\S~\ref{sec:signal}), prompting methods (\S~\ref{sec:prompt}), communication strategies (\S~\ref{sec:message}), LVLM selection (\S~\ref{sec:lvlm}), and modular design approaches (\S~\ref{sec:m3}). We investigate the framework's performance under heterogeneous agent configurations~\cite{gao2025stamp, lu2024extensible} (\S~\ref{sec:hetero}). Finally, we display some visualization results and analysis in \S~\ref{sec:vis}.

\subsection{Experimental Setup}
\label{sec:setup}

In this work we conduct closed-loop evaluations using the CARLA simulation platform. We use 10 testing scenarios in Town05 with each scenario involves two CAVs controlled by our LangCoop framework while interacting with various dynamic actors including other vehicles, pedestrians, and cyclists controlled by CARLA's traffic manager. The two CAVs are initialized at different positions within the same general vicinity. We implement V2V communication with a simulated range of 200 meters. 
For perception, each vehicle receives front-view RGB camera images at 800$\times$600 resolution.

We employ three primary evaluation metrics to assess performance comprehensively: \textbf{Driving Score (DS)}: Calculated as $\text{DS} = \text{RC} \times (1 - \text{IP})$, where RC is route completion and IP is infraction penalty. Infractions include collisions, traffic light violations, and lane invasions, each weighted according to severity. \textbf{Route Completion (RC)}: The percentage of the predefined route successfully traversed by the vehicle, measured from 0 to 100. \textbf{Time Consumed (TC)}: The total time in seconds required to complete the route or until a terminal failure. For communication efficiency assessment, we additionally track: \textbf{Transmission Bandwidth (TB)}: The average data size in KB transmitted between vehicles.

Unless otherwise specified, our baseline configuration employs GPT-4o-mini~\cite{gpt4omini} as the LVLM, utilizes a concise version of the \mcot module described in Section~\ref{sec:prompting_method}, and exchanges both front-view images (compressed JPEG) and LangPack messages between vehicles.

\subsection{Driving Signal Comparison}
\label{sec:signal}

As described in Section~\ref{sec:driving_signal}, our framework supports three distinct driving signal formulations: discrete trajectory, continuous trajectory, and direct control signals. We first compare these formulations to identify the most effective approach for subsequent experiments.

\begin{table}[htbp]
\caption{Performance comparison of different driving signal formulations. The discrete trajectory approach performs poorly while continuous trajectory and direct control signals show much stronger performance.}
\centering
\footnotesize
\begin{tabular}{lccccc}
\toprule
& \multicolumn{2}{c}{\textbf{Vehicle 1}} & \multicolumn{2}{c}{\textbf{Vehicle 2}} & \\
\cmidrule(lr){2-3}\cmidrule(lr){4-5}
\textbf{Driving Signal} & \textbf{DS} $_\red{\uparrow}$ & \textbf{RC} $_\red{\uparrow}$ & \textbf{DS} $_\red{\uparrow}$ & \textbf{RC} $_\red{\uparrow}$ & {\textbf{TC}$_{\text{(s)}}$}$_\red{\downarrow}$ \\
\midrule
Discrete Traj.             & 5.0 & 23.1 & 1.3 & 19.4 & 139.9 \\
Continuous Traj.        & 33.1 & 74.9 & \textbf{48.8} & \textbf{90.3} & \textbf{124.6}  \\
Control Signal        & \textbf{33.7} & \textbf{89.0} & 18.1 & 70.2 & 124.8  \\
\bottomrule
\end{tabular}
\label{tab:driving_signal}
\end{table}

Table~\ref{tab:driving_signal} reveals that the discrete trajectory approach performs poorly for both vehicles. This underperformance can be attributed to the poor capability of LVLMs towards discrete waypoints understandings---it is hard for zero-shot LVLMs to output discrete waypoints that are smooth and dynamically feasible. In comparison, both continuous trajectory and direct control signal approaches demonstrate better performance. 
The continuous trajectory formulation achieves better performance for Vehicle 2 (DS: 48.8, RC: 90.3), while the direct control signal approach has better performance for Vehicle 1 (DS: 33.7, RC: 89.0). The continuous trajectory approach also finish the route slightly faster than other methods.
We postulate that the strong performance of the continuous trajectory and direct control signal approaches stems from a more natural action space that better aligns with vehicle dynamics and control systems. Based on these results, we adopt the continuous trajectory approach as our default driving signal formulation for subsequent experiments for its balance of performance across both vehicles.

\subsection{Prompting Methods Comparison}
\label{sec:prompt}

Next, we evaluate three different prompting strategies to assess the impact of reasoning structure on driving performance: Naive Prompting, which directly asks the LVLM to generate driving signals without structured reasoning, Chain-of-thought (CoT), and Concise CoT. The concise CoT variation is inducing LVLMs to output a more concise description by simply adding ``Please be very concise" at the end of each prompt. 

\begin{table}[htbp]
\caption{Performance comparison of different prompting methods. The naive approach performs poorly, while both CoT approaches demonstrate strong performance.}
\centering
\footnotesize
\begin{tabular}{lccccc}
\toprule
& \multicolumn{2}{c}{\textbf{Vehicle 1}} & \multicolumn{2}{c}{\textbf{Vehicle 2}} & \\
\cmidrule(lr){2-3}\cmidrule(lr){4-5}
\textbf{Prompting} & \textbf{DS} $_\red{\uparrow}$ & \textbf{RC} $_\red{\uparrow}$ & \textbf{DS} $_\red{\uparrow}$ & \textbf{RC} $_\red{\uparrow}$ & {\textbf{TC}$_{\text{(s)}}$}$_\red{\downarrow}$ \\
\midrule
Naive & 2.7 & 23.0 & 0.7 & 21.1 & 248.7 \\
CoT       & \textbf{37.0} & \textbf{85.2} & 41.1 & 80.3 & \textbf{105.2}  \\
CoT (concise)        & 33.1 & 74.9 & \textbf{48.8} & \textbf{90.3} & 124.6  \\
\bottomrule
\end{tabular}
\label{tab:prompting}
\end{table}

Table~\ref{tab:prompting} demonstrates that the naive prompting approach performs poorly for both vehicles. This underscores the critical importance of structured reasoning for the autonomous driving task. Both CoT approaches substantially outperform the naive method, where there is no prominent performance priority bettween standard and concise CoT.
The standard CoT approach achieves the highest performance for Vehicle 1 (DS: 37.0, RC: 85.2) and completes navigation in the shortest time (105.2 seconds). Meanwhile, the concise CoT variation achieves the best performance for Vehicle 2 (DS: 48.8, RC: 90.3). 
The performance differences between standard and concise CoT prompting highlight an interesting tradeoff. The standard CoT provides more comprehensive reasoning, potentially allowing for more nuanced decision-making, while the concise version reduces computational overhead and may focus the model on the most critical aspects of the driving task. 
For subsequent experiments, we adopt the concise CoT method as our default prompting strategy, as it provides strong overall performance while maintaining computational efficiency.

\subsection{Communicative Message Comparison}
\label{sec:message}

A central aspect of our collaborative driving approach is the mechanism and content of inter-vehicle communication. We compare four different communication strategies: no collaboration (baseline), image-only sharing, LangPack-only sharing, and combined image+LangPack sharing.

\begin{table}[htbp]
\caption{Performance comparison of different communication strategies. LangPack provides substantial performance gains with minimal bandwidth usage, while the combined approach achieves the highest overall performance.}
\centering
\footnotesize
\setlength{\tabcolsep}{3.2pt}
\begin{tabular}{lccccccc}
\toprule
& \multicolumn{2}{c}{\textbf{Vehicle 1}} & \multicolumn{2}{c}{\textbf{Vehicle 2}} & \\
\cmidrule(lr){2-3}\cmidrule(lr){4-5}
\textbf{Message} & \textbf{DS} $_\red{\uparrow}$ & \textbf{RC} $_\red{\uparrow}$ & \textbf{DS} $_\red{\uparrow}$ & \textbf{RC} $_\red{\uparrow}$ & {\textbf{TC}$_{\text{(s)}}$}$_\red{\downarrow}$ & {\textbf{TB}$_{\text{(KB)}}$}$_\red{\downarrow}$ \\
\midrule
Non-collab              & 13.5 & 33.1 & 11.35 & 29.44 & 200.1 & 0 \\
Image (JPEG)         & 15.3 & 38.9 & 31.3 & 60.7 & \textbf{65.8} & 43.1  \\
LangPack         & \textbf{35.1} & \underline{71.6} & \underline{42.8} & \underline{80.1} & \underline{114.6} & 1.8 \\
Image+LangPack         & \underline{33.1} & \textbf{74.9} & \textbf{48.8} & \textbf{90.3} & 124.6 & 44.9  \\
\bottomrule
\end{tabular}
\label{tab:communication}
\end{table}

As shown in Table~\ref{tab:communication}, the non-collaborative baseline performs poorly with driving scores, which affirms the importance of multi-vehicular collaboration. The image-only strategy shows modest improvements over the non-collaborative baseline but falls significantly short of the LangPack-based methods. This suggests that raw visual data, while information-rich, may not be optimally structured for inter-vehicle understanding without additional processing. The LangPack-only approach achieves remarkable performance 
(Vehicle 1: DS 35.1, RC 71.6; Vehicle 2: DS 42.8, RC 80.1)
while requiring minimal bandwidth (1.8 KB), demonstrating the exceptional efficiency of our natural language packaging approach. This represents a bandwidth reduction of over 96\% compared to image sharing while delivering substantially better performance, The combined Image+LangPack approach achieves the highest overall performance, particularly for Vehicle 2 (DS: 48.8, RC: 90.3), but has highest bandwidth consumption (44.9 KB).

These results demonstrate that LangPack offers an exceptional balance between performance and communication efficiency, highlighting the information density and semantic richness of structured natural language representations. For bandwidth-constrained applications, LangPack-only communication provides near-optimal performance with minimal data requirements. When bandwidth constraints are less severe, the combined approach offers incremental performance improvements at the cost of substantially higher data transmission.

\begin{figure*}[t]  
    \centering
    \includegraphics[width=\textwidth]{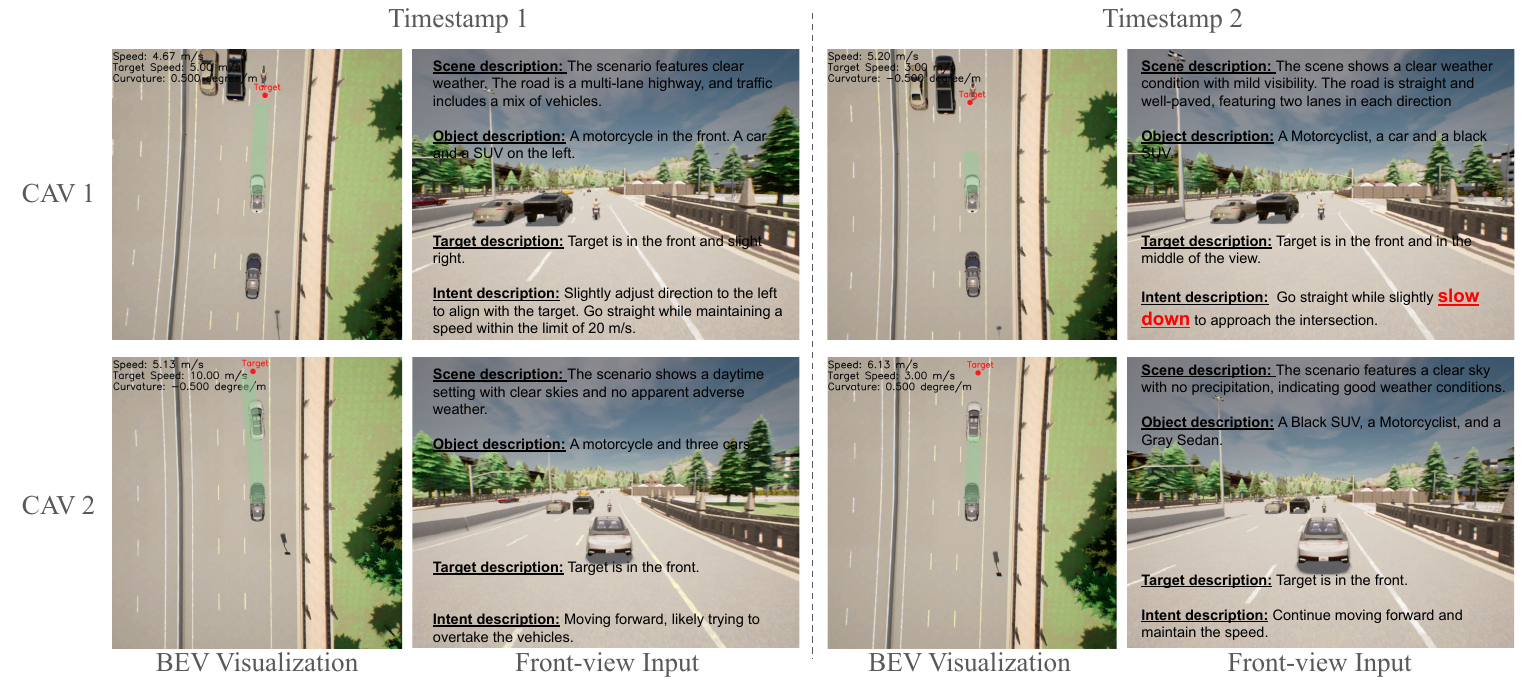} 
    \caption{Visualization of a natural-language-based collaborative driving scenario. CAV 2 slows down upon receiving the ‘slow down’ intent description from CAV 1. The context is slightly paraphrased for better visualization.}
    \label{fig:vis} 
\end{figure*}

\subsection{LVLM Performance Comparison}
\label{sec:lvlm}

The choice of LVLM significantly impacts collaborative driving performance.
We evaluate six popular vision-language models (GPT-4o, Claude-3.7 Sonnet, GPT4o-mini, Gemini Flash Lite 2.0, Qwen-2.5-VL-7B, and Llama 3.2 11B Vision Instruct) to determine their effectiveness within our framework. In the following, we refer these models as GPT-4o, Claude-3.7, GPT4o-mini, Gemini-2.0, Qwen-2.5, and Llama-3.2 respectively.
\begin{table}[htbp]
\centering
\footnotesize
\caption{Performance comparison of different LVLMs. The top-tier commercial models (GPT-4o, Claude-3.7) demonstrate the strongest performance, with GPT-4o-mini offering competitive capabilities at lower computational cost.}
\begin{tabular}{lccccc}
\toprule
& \multicolumn{2}{c}{\textbf{Vehicle 1}} & \multicolumn{2}{c}{\textbf{Vehicle 2}} & \\
\cmidrule(lr){2-3}\cmidrule(lr){4-5}
\textbf{Model} & \textbf{DS} $_\red{\uparrow}$ & \textbf{RC} $_\red{\uparrow}$ & \textbf{DS} $_\red{\uparrow}$ & \textbf{RC} $_\red{\uparrow}$ & {\textbf{TC}$_{\text{(s)}}$}$_\red{\downarrow}$ \\
\midrule
GPT-4o              & \textbf{41.3} & 70.0 & 47.7 & \underline{91.0} & 79.0  \\
Claude-3.7         & 32.0 & \underline{67.0} & \textbf{72.1} & \textbf{94.1} & 88.5  \\
GPT-4o-mini         & \underline{33.1} & \textbf{74.9} & \underline{48.8} & 90.3 & 124.6  \\
Gemini-2.0         & 12.1 & 33.7 & 25.6 & 58.0 & \textbf{46.5}  \\
Qwen-2.5            & 15.5 & 32.2 & 19.4 & 28.8 & \underline{70.7}  \\
Llama-3.2          & 11.6 & 31.1 & 19.0 & 42.2 & 102.5  \\
\bottomrule
\end{tabular}
\label{tab:vlms}
\end{table}

Table~\ref{tab:vlms} shows that GPT-4o, Claude-3.7, and GPT-4o-mini consistently outperform other options across both vehicles, suggesting these models possess superior capabilities for understanding complex driving scenes and generating appropriate driving actions in collaborative contexts. The remaining models Gemini-2.0, Qwen-2.5, and Llama-3.2 demonstrate lower performance. Interestingly, Gemini-2.0 completes routes in the shortest time (46.5 seconds), suggesting more aggressive driving behavior that may prioritize speed over safety or adherence to traffic rules. 

\subsection{Mixture Model Modular Design}
\label{sec:m3}

Our \mcot architecture enables the use of different specialized LVLMs for distinct reasoning stages. To evaluate the potential benefits of this modular approach, we implement two experimental configurations with varying model assignments for each prompting stage. In Experiment \ref{tab:m3}.A, we use Gemini-2.0 for driving scene and interactive objects description, Llama-3.2 for navigation goal and feature intent description, and use GPT4o-mini for driving signal generation. In Experiment \ref{tab:m3}.B, we use Qwen-2.5 for driving scene and interactive objects description, Llama-3.2 for navigation goal and feature intent description, and use GPT4o-mini for driving signal generation.

\begin{table}[htbp]
\caption{Performance comparison of different Mixture Model Modular (\mcot) configurations.}
\centering
\footnotesize
\begin{tabular}{lccccc}
\toprule
& \multicolumn{2}{c}{\textbf{Vehicle 1}} & \multicolumn{2}{c}{\textbf{Vehicle 2}} & \\
\cmidrule(lr){2-3}\cmidrule(lr){4-5}
\textbf{\mcot Setup} & \textbf{DS} $_\red{\uparrow}$ & \textbf{RC} $_\red{\uparrow}$ & \textbf{DS} $_\red{\uparrow}$ & \textbf{RC} $_\red{\uparrow}$ & {\textbf{TC}$_{\text{(s)}}$}$_\red{\downarrow}$ \\
\midrule
GPT4o-mini         & 33.1 & 74.9 & 48.8 & 90.3 & 124.6  \\
Exp \ref{tab:m3}.A & 31.4 & 67.9 & 37.2 & 71.3 & 144.6 \\
Exp \ref{tab:m3}.B & 35.2 & 68.5 & 42.1 & 82.6 & 119.3 \\
\bottomrule
\end{tabular}
\label{tab:m3}
\end{table}

From Table~\ref{tab:m3}, in experiments \ref{tab:m3}.A and \ref{tab:m3}.B, we observe that replacing the reasoning modules with LVLMs other than GPT4o-mini results in slightly lower but still competitive performance compared to the pure GPT4o-mini model. Given that the API costs of Gemini-2.0 and Llama-3.2 are lower than that of GPT4o-mini, these experimental results suggest that in practical scenarios with limited computational budgets, our \Mcot module supports the possibility of replacing reasoning modules with a mixture of models.

\subsection{Heterogeneous Agents Evaluation}
\label{sec:hetero}

In real-world deployments, collaborative driving systems will likely operate in environments where different vehicles utilize AI models with varying capabilities. To assess our framework’s effectiveness in such heterogeneous settings, we conduct two experiments with vehicle pairs using different LVLMs. In experiment \ref{tab:hetero}.A, the vehicles are equipped with GPT-4o-mini and Gemini-2.0, while in experiment \ref{tab:hetero}.B, they are equipped with GPT-4o-mini and Llama-3.2.

\begin{table}[htbp]
\caption{Performance in heterogeneous agent configurations where vehicles use different LVLMs.}
\centering
\scriptsize
\begin{tabular}{lllccc}
\toprule
\multicolumn{3}{c}{\textbf{Heterogeneous Setup}} & \textbf{DS} $_\red{\uparrow}$ & \textbf{RC} $_\red{\uparrow}$ & {\textbf{TC}$_{\text{(s)}}$}$_\red{\downarrow}$ \\
\midrule
\multirow{4}{*}{Exp \ref{tab:hetero}.A} & \multirow{2}{*}{Non-collab} & GPT-4o-mini & 18.2 & 56.1 & 167.3 \\
& & Gemini-2.0       & 12.6 & 61.1 & 167.3  \\
\cmidrule(lr){2-6}
& \multirow{2}{*}{Image+LangPack} & GPT-4o-mini & 59.1 & 73.2 & 126.8 \\
& & Gemini-2.0       & 45.3 & 70.2 & 126.8  \\
\midrule
\multirow{2}{*}{Exp \ref{tab:hetero}.B} & \multirow{2}{*}{Non-collab} & GPT-4o-mini        & 16.7 & 70.2 & 142.0  \\
& & Llama-3.2        & 11.5 & 51.0 & 142.0  \\
\cmidrule(lr){2-6}
& \multirow{2}{*}{Image+LangPack} & GPT-4o-mini        & 51.9 & 96.1 & 144.5  \\
& & Llama-3.2        & 12.6 & 40.1 & 144.5  \\
\bottomrule
\end{tabular}
\label{tab:hetero}
\end{table}

As shown in Table~\ref{tab:hetero}, collaboration improves both driving scores and route completion rates across both experiments. In experiment \ref{tab:hetero}.A, pairing GPT-4o-mini with Gemini-2.0, and in experiment \ref{tab:hetero}.B, pairing GPT-4o-mini with Llama-3.2, both vehicles benefit from the collaborative setup. This demonstrates that our framework is adaptable not only to homogeneous settings but also to heterogeneous environments.


\subsection{Visualization}
\label{sec:vis}

Figure \ref{fig:vis} displays a scenario where a leading CAV approaches an intersection and decides to slow down. After sharing its intent 'slow down' with other CAVs, the following vehicle also decides to slow down despite originally intending to continue forward. This demonstrates effective collaborative decision-making, as the follower vehicle appropriately adjusts its behavior based on the other CAV's communicated intent. The example illustrates how language-based communication enables real-time adaptive driving behaviors, enhancing overall traffic safety through multi-agent decision-level collaboration. Furthermore, this interaction highlights the practical value of our framework in translating natural language intents into concrete driving decisions across multiple autonomous vehicles. For more visualization results, please refer to our anonymous project page \href{https://xiangbogaobarry.github.io/LangCoop/}{https://xiangbogaobarry.github.io/LangCoop/}.
\section{Discussion, Limitations, and Future Work}

Our experiments with LangCoop reveal several key insights that inform future research directions:

\noindent \textbf{Advantage of Zero-shot LVLMs.}
Despite benefits of domain-specific training for LVLMs, zero-shot approaches offer clear advantages. They eliminate costly dataset collection and training while maintaining adaptability across diverse driving scenarios. Additionally, proprietary models like GPT and Gemini series cannot be fine-tuned by third parties. A zero-shot pipeline that leverages all LVLMs without domain-specific fine-tuning provides flexibility and accessibility for resource-limited institute.

\noindent \textbf{Computational and Latency Concerns.}
Regarding computational concerns, we note that LVLM efficiency is rapidly improving, and large models can generate trajectories for training more compact deployment models. Some novel dual-system designs\cite{tian2024drivevlm, mei2024continuously} may also alleviate the computational intensity. The conceptual advantages of language-based collaboration outweigh current computational demands, opening new possibilities for interpretable, efficient, and adaptable multi-agent driving systems.

\noindent \textbf{Prompting Strategies for Driving.}
We observed significant sensitivity to prompt formulation in driving contexts. For example, we observed that explicitly instructing the model to "avoid collisions" (which might seem obvious in driving) substantially improved performance. This suggests that current LVLMs may not fully internalize driving-specific common knowledge. This indicates potential for improvement through specialized prompts or fine-tuning approaches focused on autonomous driving scenarios. 


\noindent \textbf{Physical-Informed Control Integration.}
Our current implementation does not fully incorporate detailed vehicle dynamics into the planning pipeline. Future extensions could address this by integrating physical vehicle models (e.g., bicycle model).  Using techniques like quintic polynomial trajectory planning could ensure physically realizable motion while preserving the high-level reasoning capabilities of language models.

\noindent \textbf{Expanding V2X Communication Paradigms.}
While we currently focus on vehicle-to-vehicle communication, the approach naturally extends to broader V2X ecosystems~\cite{SAEJ2735_2024}. The unstructured nature of messages like Emergency Vehicle Alerts, Signal Phase and Timing, and Roadside Safety Alerts\cite{zimmer2025enhancing} aligns well with natural language representations. Future research could incorporate these additional message types, leveraging language models' ability to process diverse information streams within a unified framework.

\section{Conclusion}

This work introduces LangCoop, a novel framework that leverages natural language as a communication medium for multi-agent collaborative driving. We introduce Mixture Model Modular Chain-of-thought (M3CoT) for reasoning and the Natural Language Information Packaging (LangPack) for efficient data sharing. Extensive closed-loop experiments in simulation environments demonstrate that language-based collaboration not only reduces bandwidth requirements but also enhances driving performance and interpretability by including decision-level communication. Looking forward, further optimization of prompting strategies and deeper integration of vehicle dynamics promise to extend the capabilities of language-driven autonomous systems, marking a significant step toward safer and more efficient collaborative driving.






{
    \small
    \bibliographystyle{ieeenat_fullname}
    \bibliography{main}
}

\clearpage
\setcounter{page}{1}

\twocolumn[
\centering
\Large
\textbf{\thetitle}\\
\vspace{0.5em}Supplementary Material \\
\vspace{1.0em}
] 
\small
\begin{tcolorbox}[colback=gray!5!white, colframe=gray!75!black, 
title=Example of Natural Language Information Packaging (LangPack) -- \underline{Sent by Agent 1, Received by Agent 0}  ]

        \textbf{1.5 KB}
        
        \vspace{-0.5em}\noindent\rule{\linewidth}{1pt}\vspace{0em}

        Agent 1, located at: [2.69048, 69.03092], current speed: 4.518m/s.

        It's \textbf{scene description}:

        \noindent The driving scenario shows a clear, daytime environment with sunny weather. The road appears to be well-maintained, featuring multiple lanes in both directions. There is moderate traffic, including vehicles such as cars and a motorcycle. Trees line the roadside, indicating a suburban or semi-rural area. Overall, conditions are favorable for driving.
        
        It's \textbf{object description}:
        
        \noindent 1. Motorcycle  
        
          \noindent - Location: Center lane, mid-distance  
          
          \noindent - Status: Riding in the same direction as the vehicle 
          
          \noindent - Intent: Likely continuing straight ahead on the road.
        
        \noindent 2. Black SUV  
        
          \noindent - Location: Right lane, close to the front of the vehicle  
          
          \noindent - Status: Stationary  
          
          \noindent - Intent: Appears to be waiting or preparing to merge into the lane.
        
        \noindent 3. Red Car  
        
          \noindent - Location: Left lane, further ahead  
          
          \noindent - Status: Moving slowly, closer to the center divider 
          
          \noindent - Intent: Preparing to turn or change lanes.  

        It's \textbf{target description}:

        \noindent The target is 0.15402 meters to my right and 32.39753 meters to my front. 
        
        It's \textbf{intent description}: 
        
        \noindent 1. Check Environment: Identify the surrounding vehicles and road conditions.
        
        \noindent 2. Speed Control: Maintain a speed and adhere to the speed limit.
        
        \noindent 3. Adjust Direction: Slightly adjust to the right to align with the target location, ensuring no obstacles are in the way.
        
        \noindent 4. Avoid Collisions: The car ahead is too close, slow down while adjusting to ensure safe distance.

\end{tcolorbox}

\small
\begin{tcolorbox}[colback=gray!5!white, colframe=gray!75!black, 
title=Example of Natural Language Information Packaging (LangPack) -- \underline{Sent by Agent 0, Received by Agent 1} ]

        \textbf{1.4 KB}
        
        \vspace{-0.5em}\noindent\rule{\linewidth}{1pt}\vspace{0em}

        Agent 0, located at: [2.25215 80.50174], current speed: 5.630m/s.

        It's \textbf{scene description}:

        \noindent The scenario depicts a clear, sunny day with good visibility. The road is a multi-lane paved highway with light to moderate traffic. Several vehicles are moving in both directions, and there are trees lining the roadside, indicating a suburban or rural environment. The road condition appears smooth and well-maintained.
        
        It's \textbf{object description}:
        
        \noindent 1. Car (front center): Positioned in the right lane, traveling at a moderate speed. Intent appears to be straightforward driving.
        
        \noindent 2. Motorcycle (left center): Located in the left lane, approaching the vehicle. Likely intent is to overtake the car in the center lane.
        
        \noindent 3. Car (right lane, further ahead): Positioned in the right lane, moving in the same direction as others. Status indicates normal driving behavior.

        It's \textbf{target description}:

        \noindent The target is 0.86387 meters to by right and 36.0 meters to by front. The target is not an endpoint—continue moving forward after reaching it.
        
        It's \textbf{intent description}: 
        
        \noindent 1. Decelerate: Reduce speed to stay within the speed limit of 20 m/s.
            
        \noindent 2. Slightly Adjust Direction: Steer right towards the target (0.15402 meters to your right).
            
        \noindent 3. Monitor Traffic: Vehicles are ahead. To ensure a safe distance, slow down or change lanes if necessary.
            
        \noindent 4. Continue Forward: Maintain forward motion, adjusting as needed for further navigation.

\end{tcolorbox}

\end{document}